\documentclass[runningheads]{llncs}
\usepackage{amssymb}
\usepackage{float}
\usepackage{graphicx}
\usepackage{subfigure}
\usepackage{algorithm}
\usepackage{algorithmicx}
\usepackage[noend]{algpseudocode}
\usepackage{amsmath}
\usepackage{CJK}
\usepackage{verbatim}
\usepackage{url}
\usepackage{todonotes}
\begin{document}
\title{Warm-Start AlphaZero Self-Play \\Search Enhancements}
\titlerunning{Warm-Start AlphaZero Self-Play Search Enhancements}
%
\author{Hui Wang\and
Mike Preuss\and
Aske Plaat}
\authorrunning{H. Wang et al.}
%
\institute{Leiden Institute of Advanced Computer Science, Leiden University,\\ Leiden, the Netherlands\\
\email{h.wang.13@liacs.leidenuniv.nl}\\
\url{http://www.cs.leiden.edu}
}

\maketitle              
\begin{abstract}
Recently, AlphaZero has achieved landmark results in deep reinforcement learning, by providing a single self-play architecture that learned three different games at super human level. AlphaZero is a large and complicated system with many parameters, and success requires much compute power and fine-tuning. Reproducing results in other games is a challenge, and many researchers are looking for ways to improve results while reducing computational demands.  
AlphaZero's design is purely based on self-play and makes no use of labeled expert data or domain specific enhancements; it is designed to learn  from scratch. We propose a novel approach to deal with this cold-start problem
by employing simple search enhancements
at the beginning phase of self-play training, namely Rollout, Rapid Action Value Estimate~(RAVE) and dynamically weighted combinations of these with the neural network, and Rolling Horizon Evolutionary Algorithms (RHEA).
Our experiments indicate that most of these enhancements improve the performance of their baseline player in three different (small) board games, with especially RAVE based variants playing strongly. 

\end{abstract}
\section{Introduction}\label{intro}
The AlphaGo series of programs~\cite{silver2016mastering,silver2017mastering,silver2018general} achieve impressive super human level performance in board games. Subsequently, there is much interest among deep reinforcement learning researchers in self-play, and self-play is applied to many applications~\cite{tao2016principle,zhang2016doctors}. In self-play, \emph{Monte Carlo Tree Search}~(MCTS)~\cite{BrownePWLCRTPSC12} is used to train a deep neural network, that is then employed in tree searches, in which MCTS uses the network that it helped train in previous iterations. 

On the one hand, self-play is utilized to generate game playing records and assign game rewards for each training example automatically.
Thereafter, these examples are fed to the neural network for improving the model. No database of labeled examples is used. Self-play learns tabula rasa, from scratch. However, self-play suffers from a cold-start problem, and may also easily suffer from bias since only a very small part of the search space is used for training, and training samples in reinforcement learning are heavily correlated~\cite{silver2017mastering,MnihKSRVBGRFOPB15}.

On the other hand, the MCTS search enhances performance of the trained model by providing improved training examples. 
There has been much research into enhancements to improve MCTS \cite{BrownePWLCRTPSC12,plaat2020learning}, but to the best of our knowledge, few of these are used in Alphazero-like self-play, which we find surprising, given the large computational demands of self-play and the cold-start and bias problems.

This may be because AlphaZero-like self-play is still young. Another reason could be that the original AlphaGo paper~\cite{silver2016mastering} remarks about AMAF and RAVE~\cite{gelly2007combining}, two of the best known MCTS enhancements, that "AlphaGo does not employ the \emph{all-moves-as-first}~(AMAF) or \emph{rapid action value estimation}~(RAVE) heuristics used in the majority of Monte Carlo Go programs; when using policy networks as prior knowledge, these biased heuristics do not appear to give any additional benefit". Our experiments indicate otherwise, and we believe there is merit in exploring warm-starting MCTS in an AlphaZero-like self-play setting. 

We agree that when the policy network is well trained, then heuristics may not provide significant added benefit. However, when this policy network has not been well trained, especially at the beginning of the training, the neural network provides approximately random values for MCTS, which can lead to bad performance or biased training. The MCTS enhancements or specialized evolutionary algorithms such as \emph{Rolling Horizon Evolutionary Algorithms}~(RHEA) may benefit the searcher by compensating the weakness of the early neural network, providing better training examples at the start of iterative training for self-play, and quicker learning. Therefore, in this work, we first test the possibility of MCTS enhancements and RHEA for improving self-play, and then choose MCTS enhancements to do full scale experiments, the results show that MCTS with warm-start enhancements in the start period of AlphaZero-like self-play  improve iterative training with tests on 3 different regular board games, using an AlphaZero re-implementation~\cite{surag2018}.

Our main contributions can be summarized as follows:

\begin{enumerate}
\item We test MCTS enhancements and RHEA, and then choose warm-start enhancements~(Rollout, RAVE and their combinations) to improve MCTS in the start phase of iterative training to enhance AlphaZero-like self-play. Experimental results show that in all 3 tested games, the enhancements can achieve significantly higher Elo ratings, indicating that warm-start enhancements can improve AlphaZero-like self-play.

\item In our experiments, a weighted combination of Rollout and RAVE with a value from the neural network always achieves better performance, suggesting also for how  many iterations to enable the warm-start enhancement.

\end{enumerate}

The paper is structured as follows. 
After giving an overview of the most relevant literature in Sect.\,\ref{sec:relatedwork}, we describe the 
test games in Sect.\,\ref{sec:testedgames}. Thereafter, we describe the AlphaZero-like self-play algorithm in Sect.\,\ref{sec:alphazero}. Before the full length experiments in Sect.\,\ref{sec:full-exp}, an orientation experiment is performed in Sect.\,\ref{sec:RHEA}. Finally, we conclude our paper and discuss future work.

\section{Related Work}\label{sec:relatedwork}

Since MCTS was created~\cite{coulom2006efficient}, many variants have been studied~\cite{BrownePWLCRTPSC12,ruijl2014combining}, especially in  games~\cite{chaslot2008monte}. In addition, enhancements such as RAVE and AMAF have been created to improve MCTS~\cite{gelly2007combining,gelly2011monte}. Specifically, \cite{gelly2011monte} can be regarded as one of the early prologues of the AlphaGo series, in the sense that it combines online search~(MCTS with enhancements like RAVE ) and offline knowledge~(table based model) in playing small board Go. 

In self-play, the large number of parameters in the deep network as well as the large number of hyper-parameters~(see Table~\ref{defaulttab}) are a black-box that precludes understanding. The high decision accuracy of deep learning, however, is undeniable~\cite{schmidhuber61deep}, as the results in Go (and many other applications) have shown~\cite{clark2015training}. After AlphaGo Zero~\cite{silver2017mastering}, which uses an MCTS searcher for training a neural network model in a self-play loop, the role of self-play has become more and more important. The neural network has two heads: a policy head and a value head, aimed at learning the best next move, and the assessment of the current board state, respectively. 

Earlier works on self-play in reinforcement learning are~\cite{tesauro1995temporal,heinz2000new,wiering2010self,van2013reinforcement,runarsson2005coevolution}. An overview is provided in~\cite{plaat2020learning}. For instance, \cite{tesauro1995temporal,wiering2010self} compared self-play and using an expert to play backgammon with temporal difference learning. \cite{runarsson2005coevolution} studied co-evolution versus self-play temporal difference learning for acquiring position evaluation in small board Go. All these works suggest promising results for self-play. 

More recently, \cite{wang2018monte} assessed the potential of classical Q-learning by introducing Monte Carlo Search enhancement to improve training examples efficiency. \cite{wu2019accelerating} uses  domain-specific features and optimizations, but still starts from random initialization and makes no use of outside strategic knowledge or preexisting data, that can accelerate the AlphaZero-like self-play. 

However, to the best of our knowledge there is no further study on applying MCTS enhancements in AlphaZero-like self-play despite the existence of many practical and powerful enhancements. 

\section{Tested Games}\label{sec:testedgames}

\begin{figure}[t]
\centering
\subfigure[6$\times$6 Othello]{\label{fig:subfiggames:66othello}
\includegraphics[width=0.27\textwidth]{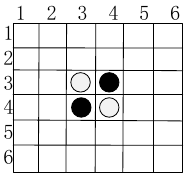}}
\hfill
\subfigure[6$\times$6 Connect Four]{\label{fig:subfiggames:66connect4}
\includegraphics[width=0.27\textwidth]{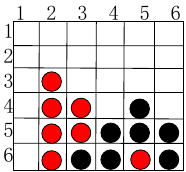}}
\hfill
\subfigure[6$\times$6 Gobang]{\label{fig:subfiggames:66Gobang}
\includegraphics[width=0.27\textwidth]{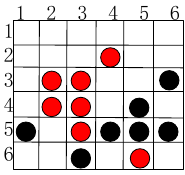}}
\caption{Starting position for Othello, example positions for Connect Four and Gobang}
\label{fig:subfiggames} 
\end{figure}

In our experiments, we use the games Othello~\cite{IwataK94}, Connect Four~\cite{Allis88} and Gobang~\cite{reisch1980gobang} with 6$\times$6 board size. All of these are two-player games. 
In Othello, any opponent's color pieces that are in a straight line and bounded by the piece just placed and another piece of the current player's are flipped to the current player's color. While the last legal position is filled, the player who has more pieces wins the game. Fig.~\ref{fig:subfiggames:66othello} show the start configurations for Othello. Connect Four is a connection game. Players take turns dropping their own pieces from the top into a vertically suspended grid. The pieces fall straight down and occupy the lowest  position within the column. The player who first forms a horizontal, vertical, or diagonal line of four  pieces wins the game. Fig.~\ref{fig:subfiggames:66connect4} is a game termination example for 6$\times$6 Connect Four where the red player wins the game. Gobang is another connection game that  is traditionally played with Go pieces on a Go board. 
 Players alternate turns, placing a stone of their color on an empty position. The winner is the first player to form an unbroken chain of 4 stones horizontally, vertically, or diagonally. Fig.~\ref{fig:subfiggames:66Gobang} is a termination example for 6$\times$6 Gobang where the black player wins the game with 4 stones in a line.

There is a wealth of research on implementing game-playing programs for these three games, using very different methods. For example, Buro created Logistello~\cite{buro1997othello} to play Othello using logistic regression. Chong et al.\ described the evolution of neural networks for learning to play Othello~\cite{ChongTW05}. Thill et al.\ applied temporal difference learning to play Connect Four~\cite{thill2014temporal}. Zhang et al.\ designed evaluation functions for Gobang~\cite{zhang2012design}. Moreover, Banerjee et al.\ tested knowledge transfer in General Game Playing on small games including 4$\times$4 Othello~\cite{banerjee2007general}. Wang et al.\ assessed the potential of classical Q-learning based on small games including 4$\times$4 Connect Four~\cite{wang2018assessing}. Varying the board size allows us to reduce or increase the computational complexity of these games.
In our experiments, we use AlphaZero-like  learning~\cite{wang2019alternative}. 

\section{AlphaZero-like Self-play Algorithms}\label{sec:alphazero}
\subsection{The Algorithm Framework}\label{a0gintroduction}
Following~\cite{silver2017mastering,silver2018general}, the basic structure of AlphaZero-like self-play is an iteration over three different stages (see Algorithm~\ref{alg:a0g}). 
\begin{algorithm*}[bth!]
\caption{AlphaZero-like Self-play Algorithm}
\label{alg:a0g}
\begin{algorithmic}[1]
\footnotesize
\Function{AlphaZeroGeneralwithEnhancements}{}
\State Initialize $f_\theta$ with random weights; Initialize retrain buffer $D$ with capacity $N$ 
\For{iteration=1, $\dots$,$I^\prime$, $\dots$, $I$}
\Comment play curriculum of $I$ tournaments
\For{episode=1,$\dots$, $E$}\Comment{stage 1, 
play tournament of $E$ games}
\For{t=1, $\dots$, $T^\prime$, $\dots$, $T$} 
\Comment play game of $T$ moves

\State$\pi_{t} \leftarrow$ \textbf{MCTS Enhancement} before $I^\prime$ or \textbf{MCTS} after $I^\prime$ iteration\label{linewithorwithoutenhancement}
\State $a_t=$randomly select on $\pi_t$ before $T^\prime$ or $\arg\max_a(\pi_t)$ after $T^\prime$ step
\State executeAction($s_t$, $a_t$)
\EndFor
\State Store every $(s_t,a_t,z_t)$ with game outcome $z_t$ ~($t\in[1,T]$) in $D$
\EndFor
\State Randomly sample  minibatch of examples~($s_j$, $\pi_j$, $z_j$) from $D$ \Comment{stage 2}
\State Train $f_{\theta^\prime}\leftarrow f_\theta$ 

\State $f_\theta=f_{\theta^\prime}$ if $f_{\theta^\prime}$ is better than $f_\theta$ using \textbf{MCTS} mini-tournament \Comment{stage 3}
\EndFor
\State \Return $f_\theta$;
\EndFunction
\end{algorithmic}
\end{algorithm*}

The first stage is a \textbf{self-play} tournament. The computer  plays several games against itself in order to generate example data for further training. In each step of a game~(episode), the player runs MCTS~(or one of the MCTS enhancements before \emph{I'} iteration) to obtain, for each move, an enhanced policy $\pi$ based on the probability $\textbf{p}$ provided by the policy network $f_\theta$. We now introduce the hyper-parameters, and the abbreviation that we use in this paper (see Table~\ref{defaulttab}). In MCTS, hyper-parameter $C_p$ is used to balance exploration and exploitation of game tree search, and we  abbreviate it to {\em c}. Hyper-parameter \emph{m} is the number of times to run down from the root for building the game tree, where the parameterized network $f_\theta$ provides the value~($v$) of the states for MCTS. For the actual (self-)play, from \emph{T'} steps on, the player always chooses the best move according to $\pi$. Before that, the player always chooses a random move based on the probability distribution of $\pi$. After finishing the games, the new examples are normalized as a form of ($s_t, \pi_t, z_t$) and stored in \emph{D}. 

The second stage consists of \textbf{neural network training},  using data from the self-play tournament. Training lasts for several epochs. In each epoch~(\emph{ep}), training examples are divided into several small batches~\cite{ioffe2015batch} according to the specific batch size~(\emph{bs}). The neural network is trained to minimize~\cite{kingma2014adam} the value of the \emph{loss function} which sums up the mean-squared error between predicted outcome and real outcome and the cross-entropy losses between $\textbf{p}$ and $\pi$ with a learning rate~(\emph{lr}) and dropout~(\emph{d}). Dropout is used as probability to randomly ignore some nodes of the hidden layer in order to avoid overfitting~\cite{srivastava2014dropout}.

The last stage is the \textbf{arena comparison}, in which  the newly trained neural network model~($f_\theta^\prime$) is run against the previous neural network model~($f_\theta$). The better model is adopted for the next iteration. In order to achieve this, $f_{\theta^\prime}$ and $f_\theta$ play against each other for $n$ games. If $f_{\theta^\prime}$ wins more than a fraction of $u$ games, it is replacing the previous best $f_\theta$. Otherwise, $f_{\theta^\prime}$ is rejected and $f_\theta$ is kept as current best model. Compared with AlphaGo Zero, AlphaZero does not entail the arena comparison stage anymore. However, we keep this stage for making sure that we can safely recognize improvements. 

\begin{algorithm*}[bth!]
\caption{Neural Network Based MCTS}
\label{alg:mcts}
\begin{algorithmic}[1]
\footnotesize
\Function{MCTS}{$s, f_\theta$}
\State Search(s)
\State $\pi_s\leftarrow$normalize($Q(s,\cdot)$)\label{line:getpolicy}
\State \Return $\pi_s$ 
\EndFunction
\Function{Search}{$s$}
\State Return game end result if $s$ is a terminal state
\If{$s$ is not in the Tree}
\State Add $s$ to the Tree, initialize $Q(s, \cdot)$ and $N(s, \cdot)$ to 0
\State Get $P(s, \cdot)$ and $v(s)$ by looking up $f_\theta(s)$ \label{mctsvaluereplace}
\State \Return $v(s)$
\Else
\State Select an action $a$  with highest UCT value\label{line:uct}
\State $s^\prime\leftarrow$getNextState($s$, $a$)
\State $v\leftarrow$Search($s\prime$)
\State $Q(s,a)\leftarrow\frac{N(s,a)*Q(s,a)+v}{N(s,a)+1}$
\State $N(s,a)\leftarrow N(s,a)+1$ \label{line:update}
\EndIf
\State \Return $v$;
\EndFunction
\end{algorithmic}
\end{algorithm*}

\subsection{MCTS}\label{subsec:mctsnnt}
In self-play, MCTS is used to generate high quality examples for training the neural network. A recursive MCTS pseudo code is given in Algorithm~\ref{alg:mcts}. 
For each search, the value from the value head of the neural network is returned (or the game termination reward, if the game terminates). During the search, for each visit of a non-leaf node, the action with the highest P-UCT value is selected to investigate next~\cite{silver2017mastering,rosin2011multi}. 
After the search, the average win rate value $Q(s,a)$ and visit count $N(s,a)$ in the followed trajectory are updated correspondingly. The P-UCT formula that is used is as follows (with $c$ as constant weight that balances exploitation and exploration): 
\begin{equation}
   U(s,a) = Q(s,a) + c*P(s,a)\frac{\sqrt{N(s,\cdot)}}{N(s,a)+1}
\end{equation}

In the whole training iterations~(including the first \emph{I'} iterations), the \textbf{Baseline} player always runs neural network based MCTS~(i.e line~\ref{linewithorwithoutenhancement} in Algorithm~\ref{alg:a0g} is simply replaced by $\pi_{t} \leftarrow \textbf{MCTS}$).

\subsection{MCTS Enhancements}\label{subsec:mctsenhancements}
In this paper, we introduce 2 individual enhancements and 3 combinations to improve neural network training based on MCTS~(Algorithm~\ref{alg:mcts}).

\noindent \textbf{Rollout}  Algorithm~\ref{alg:mcts} uses the value from the value network as return value at leaf nodes. However, if the neural network is not yet well trained, the values are not accurate, and even random at the start phase, which can lead to biased and slow training. Therefore, as warm-start enhancement we perform a classic MCTS random rollout to get a value that provides more meaningful information. We thus simply add a random rollout function which returns a terminal value after line~\ref{mctsvaluereplace} in Algorithm~\ref{alg:mcts}, written as \textit{Get result v(s) by performing random rollout until the game ends}.\footnote{In contrast to AlphaGo~\cite{silver2016mastering}, where random rollouts were mixed in with all value-lookups, in our scheme they  replace the network lookup at the start of the training.}

\noindent \textbf{RAVE} is a well-studied enhancement for improving the cold-start of MCTS in games like Go~(for details see~\cite{gelly2007combining}). The same idea can be applied to other domains where the playout-sequence can be transposed. Standard MCTS only updates the $(s, a)$-pair that has been visited. The RAVE enhancement extends this rule to any action $a$ that appears in the sub-sequence, thereby rapidly collecting more statistics in an off-policy fashion. The idea to perform RAVE at startup is adapted from AMAF in the game of Go~\cite{gelly2007combining}. The main pseudo code of RAVE is similar to Algorithm~\ref{alg:mcts}, the  differences are in line~\ref{line:getpolicy}, line~\ref{line:uct} and line~\ref{line:update}. For RAVE, in line~\ref{line:getpolicy}, policy $\pi_s$ is normalized based on $Q_{rave}(s,\cdot)$. In line~\ref{line:uct}, the action $a$ with highest $UCT_{rave}$ value, which is computed based on Equation~\ref{equationUCTrave},  is selected. After line~\ref{line:update}, the idea of AMAF is applied to update $N_{rave}$ and $Q_{rave}$, which are written as: $N_{rave}(s_{t_1},a_{t_2})\leftarrow N_{rave}(s_{t_1},a_{t_2})+1$, $Q_{rave}(s_{t_1},a_{t_2})\leftarrow\frac{N_{rave}(s_{t_1},a_{t_2})*Q_{rave}(s_{t_1},a_{t_2})+v}{N_{rave}(s_{t_1},a_{t_2})+1}$, where $s_{t_1}\in VisitedPath$, and $a_{t_2}\in A(s_{t_1})$, and for $\forall t < t_2, a_t\neq a_{t_2}$. More specifically, under state $s_t$, in the visited path, a state $s_{t_1}$, all legal actions $a_{t_2}$ of $s_{t_1}$  that appear in its sub-sequence~($t\leq t_1<t_2$) are considered as a $(s_{t_1}, a_{t_2})$ tuple to update their $Q_{rave}$ and $N_{rave}$.


\begin{equation}\label{equationUCTrave}
 UCT_{rave}(s,a)=(1-\beta)* U(s,a) + \beta*U_{rave}(s,a)
 \end{equation}
 where 
 \begin{equation}
 U_{rave}(s,a) = Q_{rave}(s,a) + c*P(s,a)\frac{\sqrt{N_{rave}(s,\cdot)}}{N_{rave}(s,a)+1},
 \end{equation}
 and 
 \begin{equation}
 \beta=\sqrt{\frac{equivalence}{3*N(s,\cdot)+equivalence}}
 \end{equation}
 Usually, the value of equivalence is set to the number of MCTS simulations~(i.e $m$), as is also the case in our following experiments.

\noindent \textbf{RoRa} Based on Rollout and Rave enhancement, the first combination is to simply add the random rollout to enhance RAVE. 

\noindent \textbf{WRo} As the neural network model is getting better, we introduce a weighted sum of rollout value and the value network as the return value. In our experiments, $v(s)$ is computed as follows:
\begin{equation}
v(s)=(1-weight)*v_{network}+ weight*v_{rollout}
\end{equation}

\noindent \textbf{WRoRa} In addition, we also employ a weighted sum to combine the value a neural network and the value of RoRa. In our experiments, weight $weight$ is related to the current iteration number $i, i\in[0,I']$.  $v(s)$ is computed as follows:
\begin{equation}
v(s)=(1-weight)v_{network}+ weight*v_{rora}
\end{equation}
where
\begin{equation}
    weight=1-\frac{i}{I'}
\end{equation}

\section{Orientation Experiment: MCTS(RAVE) vs. RHEA}\label{sec:RHEA}

Before running full scale experiments on warm-start self-play that take days to weeks, we consider other possibilities for methods that could be used instead of MCTS variants. Justesen et al.~\cite{justesen2017} have recently shown that depending on the type of game that is played, RHEA can actually outperform MCTS variants also on adversarial games. Especially for long games, RHEA seems to be strong because MCTS is not able to reach a good tree/opening sequence coverage.

The general idea of RHEA has been conceived by Perez et al.~\cite{perez2013} and is simple: they directly optimize an action sequence for the next actions and apply the first action of the best found sequence for every move. Originally, this has been applied to one-player settings only, but recently different approaches have been tried also for adversarial games, as the co-evolutionary variant of Liu et al.~\cite{Liu2016} that shows to be competitive in 2~player competitions~\cite{Gaina2018}.
The current state of RHEA is documented in \cite{gaina2020rolling}, where a large number of variants, operators and parameter settings is listed. No one-beats-all variant is known at this moment.

Generally, the horizon (number of actions in the planned sequence) is often much too short to reach the end of the game. In this case, either a value function is used to assess the last reached state, or a rollout is added. For adversarial games, opponent moves are either co-evolved, or also played randomly. We do the latter, with a horizon size of $10$. 
In preliminary experiments, we found that a number of $100$ rollouts is already working well for MCTS on our problems, thus we also applied this for the RHEA. In order to use these $100$ rollouts well, we employ a population of only 10 individuals, using only cloning+mutation (no crossover) and a (10+1) truncation selection (the worst individual from 10 parents and 1 offspring is removed). The mutation rate is set to $0.2$ per action in the sequence. However, parameters are not sensitive, except rollouts. RHEA already works with 50 rollouts, albeit worse than with 100. As our rollouts always reach the end of the game, we usually get back $Q_i(as) = \{1,-1\}$ for the $i$-th rollout for the action sequence $as$, meaning we win or lose. Counting the number of steps until this happens $h$, we compute the fitness of an individual to  
$Q(as) = \frac{\sum^{n}_{i=1} Q_i(as)/h}{n}$ over multiple rollouts, thereby rewarding quick wins and slow losses. We choose $n=2$ (rollouts per individual) as it seems to perform a bit more stable than $n=1$. We thus evaluate 50 individuals per run.

In our comparison experiment, we pit a random player, MCTS, RAVE (both without neural network support but a standard random rollout), and RHEA against each other with 500 repetitions over all three games, with 100 rollouts per run for all methods. The results are shown in Table~\ref{rheatab}.

\begin{table}[tbh!]
\centering\hspace*{-2.3em}
\footnotesize
\caption{Comparison of random player, MCTS, Rave, and RHEA on the three games, win rates in percent (column vs row), 500 repetitions each.}\label{rheatab}
\begin{tabular}{|r|r|r|r|r|r|r|r|r|r|r|r|r|}
\hline
  & \multicolumn{4}{c|}{Gobang} & \multicolumn{4}{c|}{Connect Four} & \multicolumn{4}{c|}{Othello} \\
\hline
adv& rand & mcts & rave & rhea & rand & mcts & rave & rhea & rand & mcts & rave & rhea\\
\hline
random	& &	97.0 &  100.0 &   90.0 & & 99.6 & 100.0 & 80.0 &  &	98.50 & 	98.0 & 	48.0\\

mcts &	3.0 & & 89.4 & 34.0 &	0.4 & &  73.0 & 3.0 & 1.4 & & 46.0 & 1.0\\

rave &	0.0 & 10.6 & & 	17.0 & 0.0 & 27.0 & & 4.0 &	2.0 & 54.0 & & 5.0\\

rhea & 	10.0 & 	66.0 & 	83.0 & & 20.0 & 97.0 & 96.0 & & 52.0 & 	99.0 & 95.0 & \\	

\hline

\end{tabular}
\end{table}

 The results indicate that in nearly all cases, RAVE is better than MCTS is better than RHEA is better than random, according to a binomial test at a significance level of $5\%$. Only for Othello, RHEA does not convincingly beat the random player. We can conclude from these results that RHEA is no suitable alternative in our case. The reason for this may be that the games are rather short so that we always reach the end, providing good conditions for MCTS and even more so for RAVE that more aggressively summarizes rollout information. Besides, start sequence planning is certainly harder for Othello where a single move can change large parts of the board.

\section{Full Length Experiment}\label{sec:full-exp}

Taking into account the results of the comparison of standard MCTS/RAVE
and RHEA at small scale, we now focus on the previously defined neural network based MCTS and its enhancements and run them over the full scale training.

\subsection{Experiment Setup}\label{sec:setup}

For all 3 tested games and all experimental training runs based on Algorithm~\ref{alg:a0g}, we set parameters values in Table~\ref{defaulttab}. Since tuning \emph{I'} requires enormous computation resources, we set the value to 5 based on an initial experiment test, which means that for each self-play training, only the first 5 iterations will use one of the warm-start enhancements, after that, there will be only the MCTS in Algorithm~\ref{alg:mcts}. Other parameter values are set based on~\cite{wang2019hyper,wang2020analysis}.

Our experiments are run on a GPU-machine with 2x Xeon Gold 6128 CPU at 2.6GHz, 12 core, 384GB RAM
and 4x NVIDIA PNY GeForce RTX 2080TI. We use small versions of games (6$\times$6) in order to perform a sufficiently high number of computationally demanding experiments. Shown are graphs with errorbars of 8 runs, of 100 iterations of self-play. Each single run takes 1 to 2 days.

\begin{table}[bht!]
\centering\hspace*{-2.3em}
\footnotesize
\caption{Default Parameter Setting}\label{defaulttab}
\begin{tabular}{|l|l|l|l|l|l|}
\hline
Para& Description & Value&Para& Description & Value\\
\hline
\emph{I}	&number of iteration		&100 &\emph{rs}	& number of retrain iteration		&20\\

\emph{I'}	&iteration threshold	&5	&\emph{ep}	& number of epoch	&10	\\

\emph{E}    &number of episode		&50&\emph{bs}	& batch size	&64\\

\emph{T'}	&step threshold		&15&\emph{lr}	& learning rate	&	0.005\\

\emph{m}	&MCTS simulation times	&100&\emph{d}    & dropout probability &0.3\\

\emph{c}    &weight in UCT		&1.0&\emph{n}   & number of comparison games	&40\\

\emph{u}	& update threshold		&0.6&&&\\
\hline
\end{tabular}
\end{table}

\subsection{Results}\label{sec:experimentresults}
After training, we collect 8 repetitions for all 6 categories players. Therefore we obtain 49 players in total~(a Random player is included for comparison). In a full round robin tournament, every 2 of these 49 players are set to pit against each other for 20 matches on 3 different board games~(Gobang, Connect Four and Othello). The Elo ratings are calculated based on the competition results using the same Bayesian Elo computation~\cite{coulom2008whole} as AlphaGo papers.

\begin{figure}[!b]
\centering
\hspace*{-1.5em}
\subfigure[6$\times$6 Gobang]{\label{fig:gobangtournamentresults}
\includegraphics[width=0.49\columnwidth]{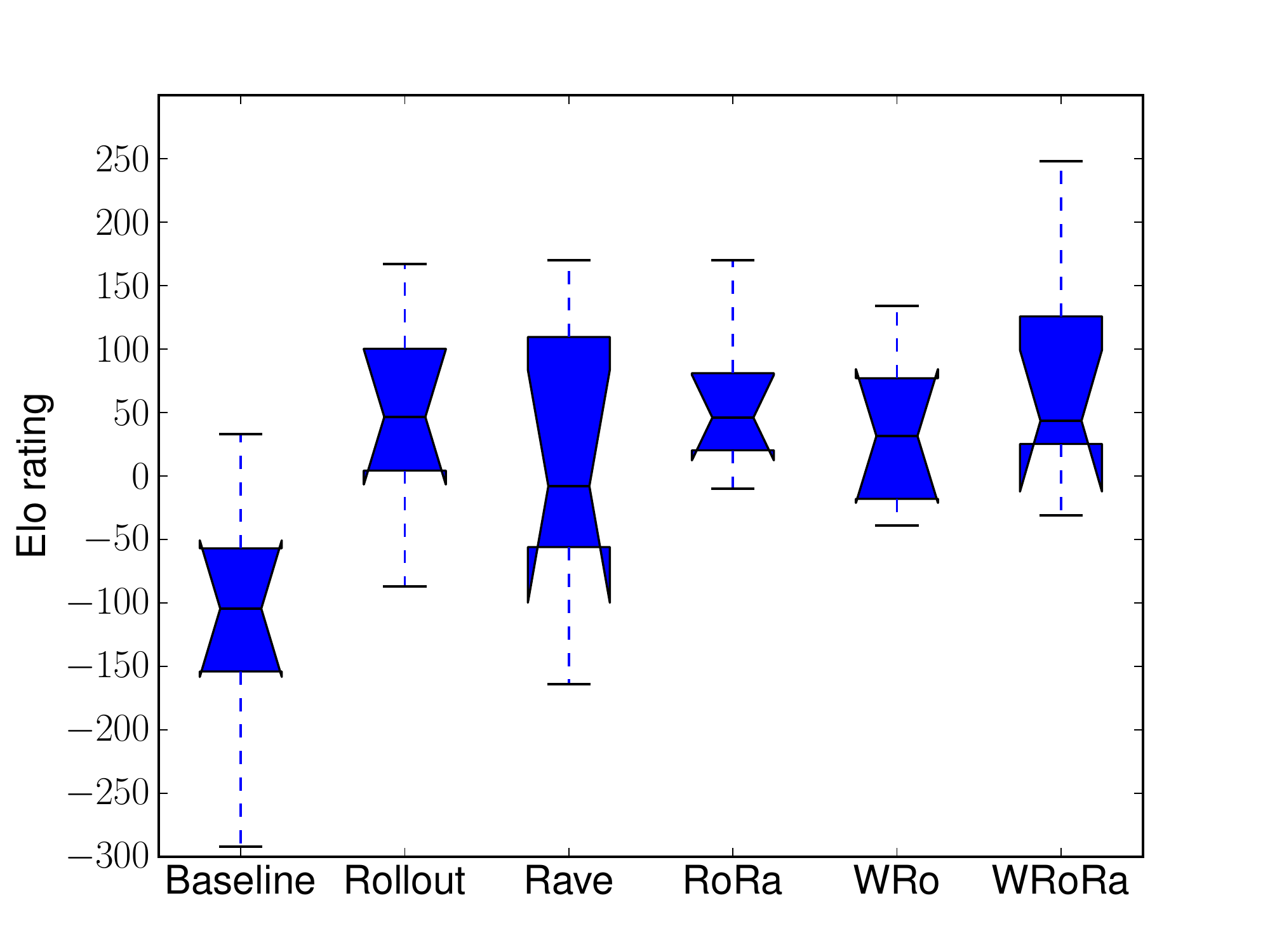}}
\hspace*{-1.8em}
\subfigure[6$\times$6 Connect Four]{\label{fig:connect4tournamentresults}
\includegraphics[width=0.49\columnwidth]{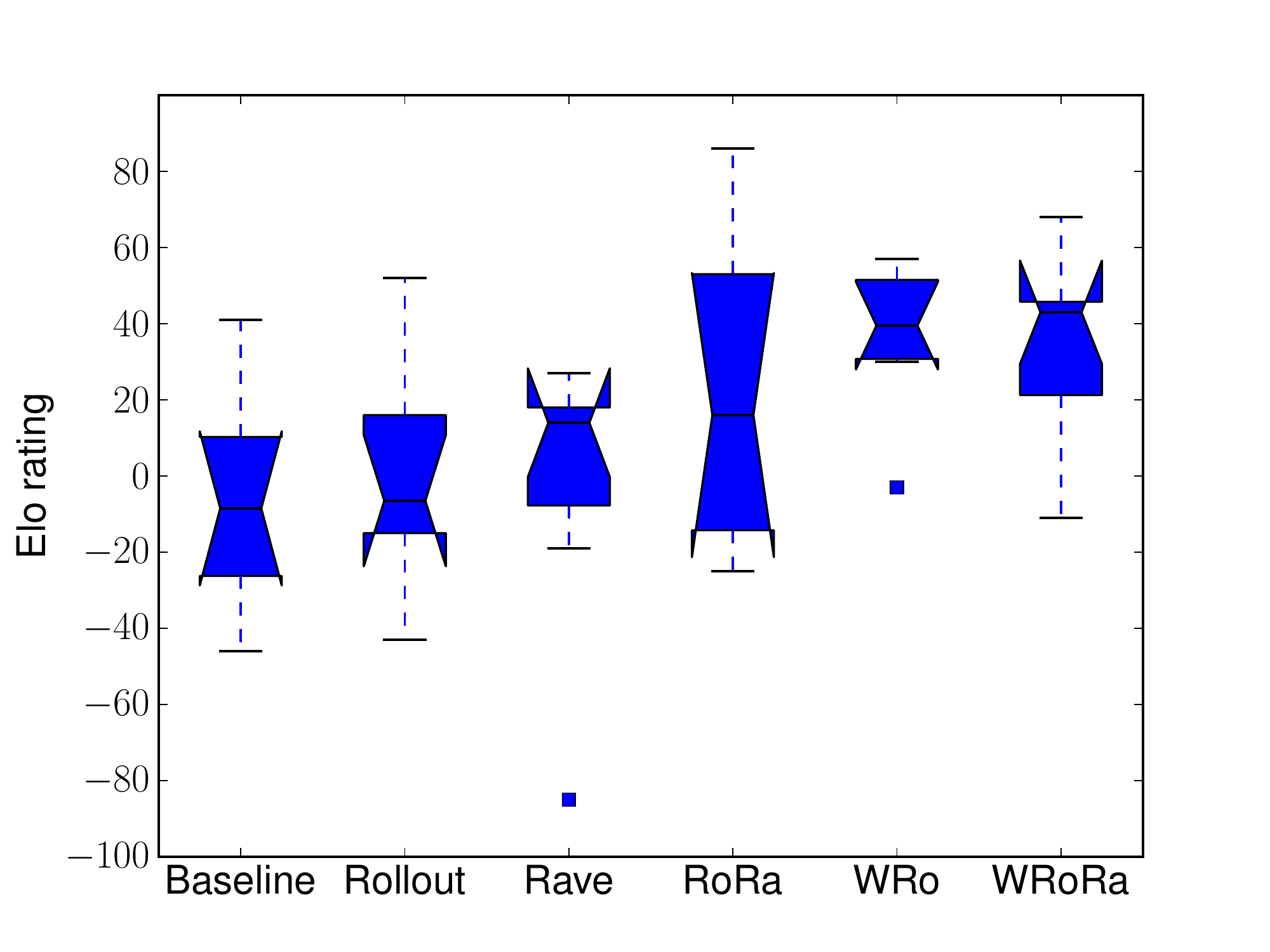}}

\caption{Tournament results for 6$\times$6 Gobang and 6$\times$6 Connect Four among \emph{Baseline}, \emph{Rollout}, \emph{Rave}, \emph{RoRa}, \emph{WRo} and \emph{WRoRa}. Training with enhancements tends to be better than baseline MCTS. 
}
\label{fig:subfigarenagobangandconnectfour} 
\end{figure}

Fig.~\ref{fig:gobangtournamentresults} displays results for training to play the 6$\times$6 Gobang game. We can clearly see that all players with the enhancement achieve higher Elo ratings than the Baseline player.
For the Baseline player, the average Elo rating is about -100. For enhancement players, the average Elo ratings are about 50, except for Rave, whose variance is larger.  Rollout players and its combinations are better than the single Rave enhancement players in terms of the average Elo. In addition, the combination of Rollout and RAVE does not achieve significant improvement of Rollout, but is better than RAVE. This indicates than the contribution of the Rollout enhancement is larger than RAVE in Gobang game.


Figure~\ref{fig:connect4tournamentresults} shows that all players with warm-start enhancement achieve higher Elo ratings in training to play the 6$\times$6 Connect Four game. In addition, we find that  comparing Rollout with WRo, a weighted sum of rollout value and neural network value achieves higher performance. Comparing Rave and WRoRa, we see the same. We conclude that in 5 iterations, for  Connect Four, enhancements that combine the value derived from the neural network
contribute more than the pure enhancement value. Interestingly, in Connect Four, the combination of Rollout and RAVE shows improvement, in contrast to Othello (next figure) where we do not see significant improvement. However, this does not apply to WRoRa, the weighted case. 

\begin{figure}[t!]
\centering
\hspace*{-1.5em}
\includegraphics[width=0.5\columnwidth]{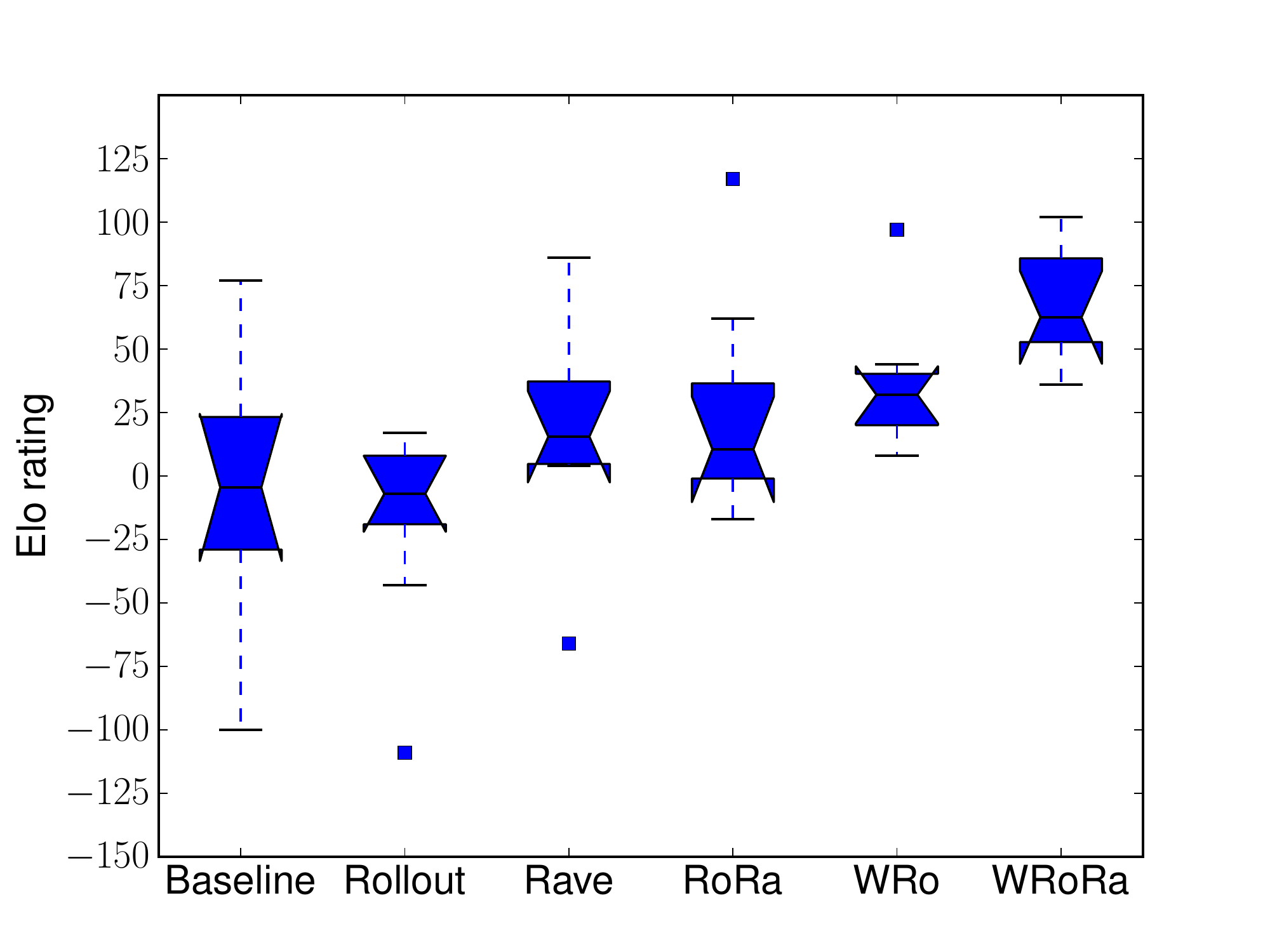}
\caption{Tournament results for 6$\times$6 Othello among \emph{Baseline}, \emph{Rollout}, \emph{Rave}, \emph{RoRa}, \emph{WRo} and \emph{WRoRa}.  Training with enhancements is mostly better than the baseline setting.}
\label{fig:othellotournamentresults}
\end{figure}

In Fig~\ref{fig:othellotournamentresults} we see that in Othello, except for Rollout which holds the similar Elo rating as Baseline setting, all other investigated enhancements are better than the Baseline. Interestingly, the enhancement with weighted sum of RoRa and neural network value achieves significant highest Elo rating. The reason that Rollout does not show much improvement could be that the rollout number is not large  enough for the game length~(6$\times$6 Othello needs 32 steps for every episode to reach the game end, other 2 games above may end up with vacant positions). In addition, Othello does not have many transposes as Gobang and Connect Four which means that RAVE can not contribute to a significant improvement. We can definitively state that the improvements of these enhancements are sensitive to the different games. In addition, for all 3 tested games, at least WRoRa achieves the best performance according to a binomial test at a significance level of $5\%$.

\section{Discussion and Conclusion}\label{sec:conclusion}
Self-play has achieved much interest due to the AlphaGo Zero results. However, self-play is currently computationally very demanding, which hinders reproducibility and experimenting for further improvements. In order to improve performance and speed up training, in this paper, we investigate the possibility of utilizing MCTS enhancements to improve AlphaZero-like self-play. We embed Rollout, RAVE and their possible combinations as enhancements at the start period of iterative self-play training. The hypothesis is, that self-play suffers from a cold-start problem, as the neural network and the MCTS statistics are initialized to random  weights and zero, and that this can be cured by prepending it with running MCTS enhancements or similar methods alone in order to train the neural network before "switching it on" for playing.

We introduce Rollout, RAVE, and combinations with network values, in order to quickly improve MCTS tree statistics before we switch to Baseline-like self-play training, and test these enhancements on 6x6 versions of Gobang, Connect Four, and Othello. 
We find that, after 100 self-play iterations, we still see the effects of the warm-start enhancements as playing strength has improved in many cases. For different games, different methods work best; there is at least one combination that performs better. It is hardly possible to explain the performance coming from the warm-start enhancements and especially to predict for which games they perform well, but there seems to be a pattern: Games that enable good static opening plans probably benefit more. For human players, it is a common strategy in Connect Four to play a middle column first as this enables many good follow-up moves. In Gobang, the situation is similar, only in 2D. It is thus harder to counter a good plan because there are so many possibilities. This could be the reason why the warm-start enhancements work so well here. For Othello, the situation is different, static openings are hardly possible, and are thus seemingly not detected.
One could hypothesize that the warm-start enhancements recover human expert knowledge in a generic way. Recently, we have seen that human knowledge is essential for mastering complex games as StarCraft~\cite{vinyals2019grandmaster}, whereas others as Go~\cite{silver2017mastering} can be learned from scratch. Re-generating human knowledge may still be an advantage, even in the latter case.

We also find that often, a single enhancement may not lead to significant improvement. There is a tendency for the enhancements that work in combination with the value of the neural network to be stronger, but that also depends on the game. 
Concluding, we can state that we find moderate performance improvements when applying warm-start enhancements and that we expect there is untapped potential for more performance gains here.

\section{Outlook}
We are not aware of other studies on warm-start enhancements of AlphaZero-like self-play. Thus, a number of interesting problems remain to be investigated.
\begin{itemize}
    \item Which enhancements will work best on which games? Does the above hypothesis hold that games with more consistent opening plans benefit more from the warm-start?
    \item When (parameter $I^\prime$) and how do we lead over from the start methods to the full AlphaZero scheme including MCTS and neural networks? If we use a weighting, how shall the weight be changed when we lead over? Linearly?
    \item There are more parameters that are critical and that could not really be explored yet due to computational cost, but this exploration may reveal important performance gains. 
    \item Other warm-start enhancements, e.g. built on variants of RHEA's or hybrids of it, shall be explored.
    \item All our current test cases are relatively small games. How does this transfer to larger games or completely different applications? 
\end{itemize}
In consequence, we would like to encourage other researchers to help exploring this approach and enable using its potential in future investigations.

\section*{Acknowledgments.} Hui Wang acknowledges financial support from the China Scholarship Council (CSC), CSC No.201706990015.
%
%
%
%

\newpage

\bibliographystyle{unsrt}
\bibliography{references}

\end{document}